\pdfoutput=1

\documentclass[11pt]{article}

\usepackage[final]{acl}

\usepackage{times}
\usepackage{latexsym}
\usepackage{graphicx}
\usepackage{booktabs}
\usepackage{multirow}

\usepackage{subcaption}

\usepackage[T1]{fontenc}

\usepackage[utf8]{inputenc}

\usepackage{microtype}

\usepackage{inconsolata}

\title{Building a Language-Learning Game for Brazilian Indigenous Languages: A Case of Study}

\author{Gustavo Polleti \\
  Universidade de S\~ao Paulo \\
  \texttt{gustavo.polleti@usp.br} 
}

\begin{document}
\maketitle
\begin{abstract}

In this paper we discuss a first attempt to build a language learning game for brazilian indigenous languages and the challenges around it. We present a design for the tool with gamification aspects. Then we describe a process to automatically generate language exercises and questions from a dependency treebank and a lexical database for Tupian languages. We discuss the limitations of our prototype highlighting ethical and practical implementation concerns. Finally, we conclude that new data gathering processes should be established in partnership with indigenous communities and oriented for educational purposes.

\end{abstract}

\section{Introduction}

Language learning games are key tools to vitalize endangered languages~\cite{thomason15, xu-etal-2022-faoi, neubig-etal-2020-summary}. LARA~\cite{akhlaghi19}, a multi language learning assistant, is an example that has been key to support actions related to endangered languages protection~\cite{rayner-wilmoth-2023-using, branislav22, zuckermann-etal-2021-lara}. Despite the necessity of language learning tools to vitalize endangered languages, they are typically restricted to high-resource languages, such as english, and require significant effort to be extended to languages with few spoken and written resources. For example, ``7000 languages''~\footnote{https://www.7000.org/}, a non-profit organization dedicated to build online courses for endangered languages, require 1 to 2 years to build a language course. As a result, despite the immediate need, language learning tools are expensive and, as they are developed today, are hard to scale to cover all the 2,680 languages in risk of being extinct by the end of this century~\cite{wurm01}. In particular, Brazil hosts approximately 270 indigenous languages, all of which are endangered. Brazil indigenous languages, collectively known as Brazilian Indigenous Languages (BILs) henceforth, are spoken by at most 30 thousand people, few of which are young children and teenagers. Brazilian indigenous communities require language learning tools that can tech their native language to portuguese speakers. Since portuguese is a low-resource language, when compared to English for example, and due to the lack of data resources on BIL, brazilian indigenous communities are underserved by current learning tools.

In this work, we describe the process of building a language learning tool for BIL, which we will refer as ``BILingo''. BILingo is a language learning game app, heavily inspired by industry leaders on language learning apps (Duolingo~\footnote{https://www.duolingo.com/} and Busuu~\footnote{https://www.busuu.com/}). We discuss in detail the challenges of building a language learning tool for BIL, such as the lack of written and phonetical resources, ethical concerns on available treebanks and databases used for exercise generation, and provide some suggestions on steps forward. We managed to build a minimal proof of concept course for Guajajara language divided in two sections. We employed dependency treebanks and a lexical database on BIL as source for exercise generation. The main contribution of this work is to present a case of study on building a language learning tool for BIL and, we hope, it will serve as an starting point for the development of an actual fully fledged language learning app that can be used to strengthen the culture of indigenous communities in Brazil.

The paper is organized as follows. Section \ref{sec:method} presents BILingo's design and describe its development process, including their data sources and exercise's format. In Section~\ref{sec:challenges} we discuss the challenges and limitations of our prototype, we analyse our processes and resources from both a practical implementation and ethical perspective. Finally, in Section~\ref{sec:conclusion} we offer concluding remarks.

\section{Method and Results} \label{sec:method}

\begin{figure*} 
    \centering
    \includegraphics[scale=0.3]{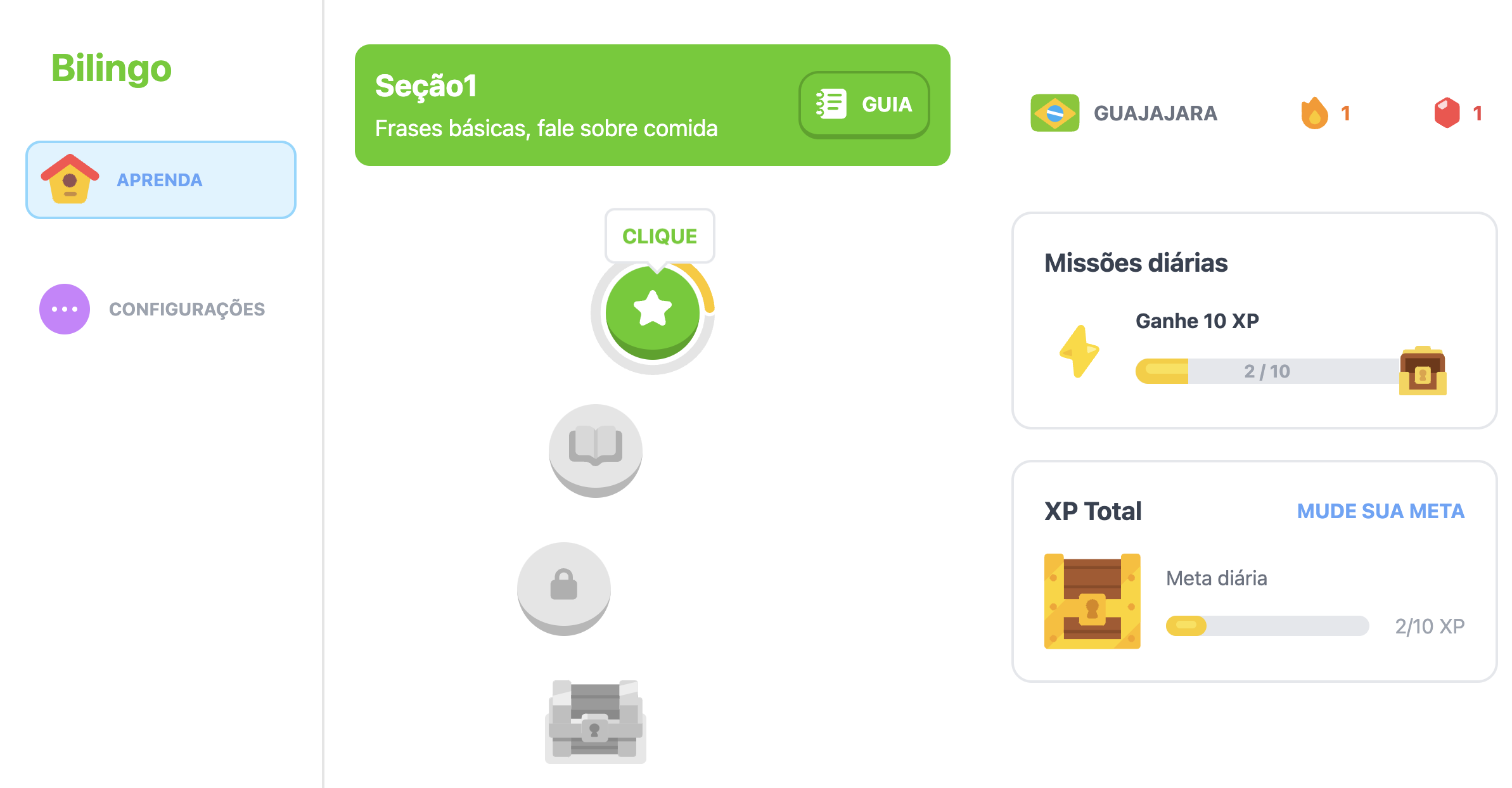}
    \caption{Landing page of BILingo. It displays a linear progress trajectory of lessons and incorporates gamification aspects such as daily quests and daily streak.}
    \label{fig:course_unit}
\end{figure*}

BILingo's design follows the gamified language learning structure found in apps available in the industry (e.g. Duolingo) and in the literacy~\cite{lightbown21, vonahn06, katinskaia-etal-2017-revita}. It has an inital course page as depicted in Figure~\ref{fig:course_unit}. The student will progress linearly in the course by completing lessons. To complete a lesson the student needs to pass a series of exercises. Every time they make a mistake, they lose a ``red gem'', if the student is out of red gems, they have to wait 5 minutes before trying a lesson again. Once the student completes a lesson, they can advance to the next one. These are gamification aspects typically found in language learning apps in the industry.

\begin{figure*}
\centering
\begin{subfigure}{.45\textwidth}
  \centering
    \includegraphics[scale=0.32]{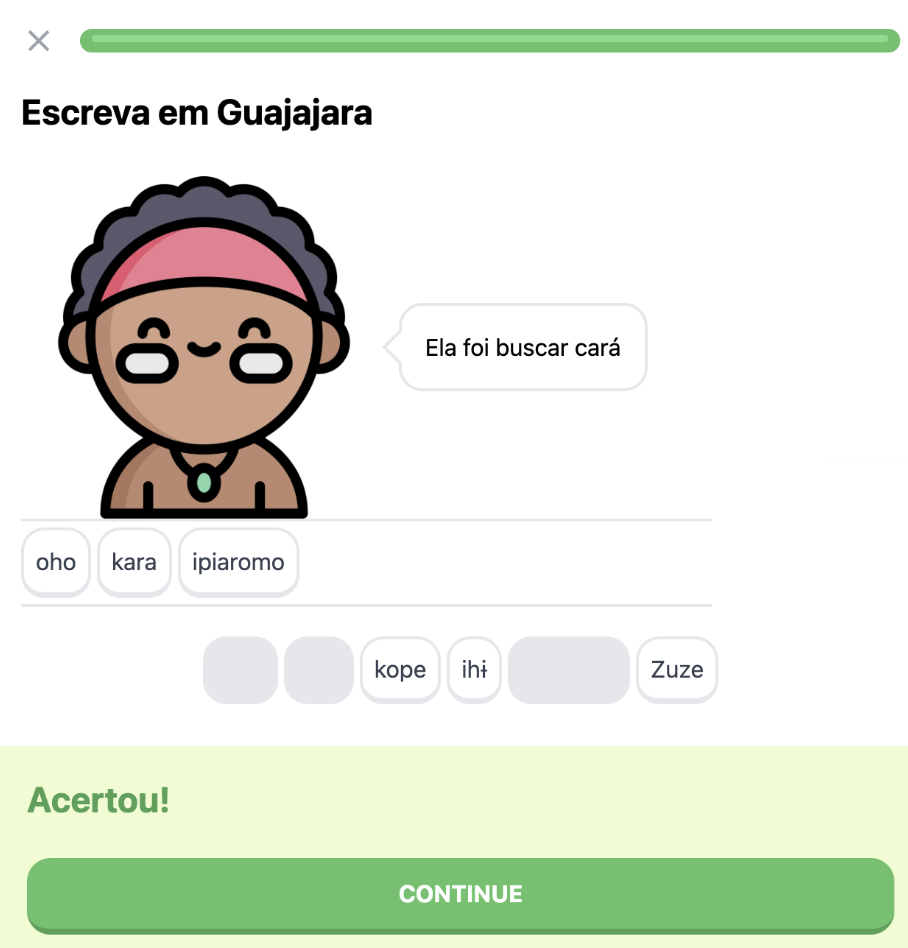}
    \caption{Translate sentence exercise type example. The student was asked to form the sentence ``She went to look for yam'' using the Guajajara word set.}
    \label{fig:exercise_ts1}
\end{subfigure}   
\hspace{5mm}
\begin{subfigure}{.45\textwidth}
  \centering
    \includegraphics[scale=0.2]{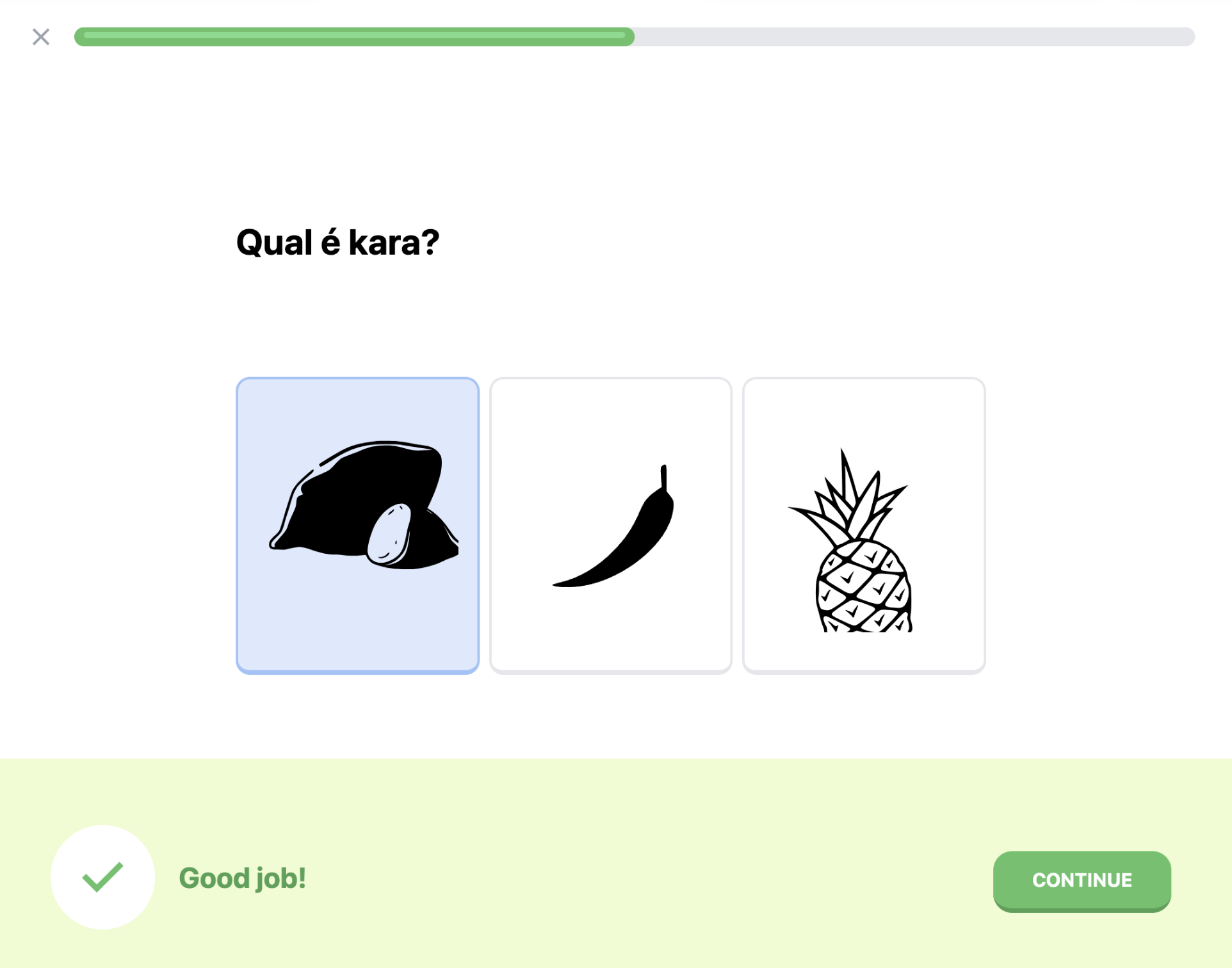}
    \caption{Concept match exercise type example. The student was asked to tell which of the images corresponded to the Guajajara word ``kara'' that means yam.}
    \label{fig:exercise_mm}
\end{subfigure}
\caption{Exercise types in our learning language tool. Depicts scenarios where the student has answered correctly.}
\label{fig:exercise_types}
\end{figure*}

When the student engages with a lesson, they can find three different types of language exercises: (1) ``translate sentence'' TS1, (2) ``translate sentence in the target language'' TS2 or (2) ``concept match'' CM, see Figure~\ref{fig:exercise_types}. In the TS1 exercise, depicted in Figure~\ref{fig:exercise_ts1}, the student is presented with a sentence in portuguese and asked to select the tokens from the Guajajara language in the correct order to form the translated sentence; TS2 is the same but the initial sentence is presented in Guajajara and the student is asked to translate it to portuguese. CM exercises present a word in Guajajara and images of possible concepts that are represented by that word, then, the student is asked to select which one of the images correspond to the given word, see Figure~\ref{fig:exercise_mm}. Popular language learning assistants, such as LARA, employ phonetical exercises that are absent in our prototype. At this point, we focused on written exercises only, with the purpose of simplify the setup and because we couldn't find any phonetical databases readily available.

Now that we presented our tool, we describe the details of its implementation. In order to build BILingo's question database for BIL, we used a simple exercise generation method based on available treebanks and lexical databases. In this work we used TuLeD~\cite{tuled22} and TuDeT~\cite{tudet22}, which are respectively a lexical database and a dependency treebank for several Tupian languages, including Guajajara. These databases compile several resources from the literature on Tupian languages and structure them into a single format suitable for analytical purposes. TuDeT treebank offers several dependency trees in the original indigenous language with some correspondent sentences in other mainstream languages, such as english, spanish, portuguese and french. For example, the token list (``oho'', ``kara'', ``ipiaromo'') corresponds to the sentence ``Ela foi buscar inhame'' in portuguese. On the other hand, TuLeD offers a lexical database with the ontological concept associated to each term. So, in our example, you will find that the word ``kara'' means ``yam''. In order to link both databases, we first conducted a search of all the concepts available in TuLeD on TuDeT sentences, so that we could tell which concepts were present in each dependency tree. Here, we considered a hit if the exact same form that appeared in each dependency tree was present in the lexical database. For example, we could tell that the sentence ``oho kara ipiaromo'' refers to the concept ``yam'' on the word ``kara''.

As we presented, BILingo consists in 3 concepts: ``course section'', ``lesson'' and ``exercise''. First, we had to determine with the available resources at hand, our unified TuLeD and TuDeT database, which topics or subjects we could cover in our prototype. To select the topics, we grouped the available dependency trees for each language by concept, and then manually inspected which concepts were suitable for building course sections. For example, concepts like ``yam'', ``pineapple'' and ``pepper'' appear in a significant number of sentences, e.g. more than 10, we can tell that ``food'' can be a good candidate as course subject. Once we selected the subjects for a given course section, we can filter only the sentences related to the listed concepts. Now we can generate CM exercises by sampling from the listed concepts within the same section. Furthermore, we can also use the dependency trees, with their correspondent translation in portuguese, to build class TS1 and TS2 exercises. To generate tokens other than the correct ones for the TS1 and TS2 exercises, we can simply select a sample set of dependency trees from the same language, shuffle their tokens and apply a random sampling. The lessons were generated by randomly sampling a predefined number of exercises, e.g. 4, from our set, always ensuring to have two of each kind. In this work, we were able to generate exercises for two course sections, each one comprising 4 lessons. Table~\ref{tab:translate_sentence_table} presents an example list of exercises generated through our process.

\begin{table*}
    \caption{BILingo example exercises.}
    \label{tab:translate_sentence_table}
    
    \centering
    \begin{tabular}{l|l|l}
    \toprule
    \textbf{Section} & \textbf{Question (translate sentence)} & \textbf{Answer} \\
    \midrule
        food
        & ela foi buscar cará
        & oho kara ipiaromo \\\hline

        food
        & a mãe de josé foi a roça para buscar carã
        & oho zuze ihi kope kara ipiaromo \\\hline

        food
        & tem abacaxi na roça de josé
        & heta nana zuze kope \\\hline

        food 
        & ele colhe cacau
        & opo?o aka?u a?e \\\hline

        animal 
        & a mulher envolveu o peixe
        & owan kuza pira a?e \\\hline

        animal 
        & foi o queixado
        & tazahu ru?u \\\hline

        animal 
        & o que foi que o queixado comeu na roça
        & ma?e tazahu u?u kope ra?e \\\hline

        animal 
        & o homem alimentou o peixe
        & opoz awa pira a?e \\
        
    \bottomrule
    \end{tabular}
\end{table*}

\section{Challenges and Limitations} \label{sec:challenges}

\begin{table}[t]
    \caption{Statistics on sources for exercise generation. For each language, we have the count of sentences that had at least a single concept associated (has\_concept) and their respective translations to portuguese (pt) and english (en).}
    \centering
    \begin{tabular}{llrrr}
    \toprule
     language & has\_concept & pt & en \\
    \midrule
    Tupinamba & True & 0 & 140 \\
    Tupinamba & False & 0 & 409 \\
    Teko & True & 0 & 19 \\
    Teko & False & 0 & 95 \\
    Munduruku & True & 22 & 22 \\
    Munduruku & False & 155 & 156 \\
    Makurap & True & 0 & 15 \\
    Makurap & False & 0 & 37 \\
    Karo & True & 0 & 260 \\
    Karo & False & 0 & 664 \\
    Kaapor & True & 0 & 58 \\
    Kaapor & False & 0 & 83 \\
    Guajajara & True & 719 & 487 \\
    Guajajara & False & 1172 & 806 \\
    Akuntsu & True & 25 & 186 \\
    Akuntsu & False & 36 & 325 \\
    \bottomrule
    \end{tabular}
    \label{tab:source_link}
\end{table}

Our prototype falls short in several aspects, from the difficulties of working with limited sources of data for exercise generation to ethical concerns, now we examine all the learnings and challenges to actually build a working system for BIL. As we discussed, our work relies on TuLeD and TuDeT as source for exercise generation. Both databases were developed by compiling several sources from the literature, so there was no structured data gathering process and, thus, the data may be seem questionable in many senses. First, we could observe that it is severely incomplete, notably when we consider coverage of dependency trees with translation to portuguese. Since many works that were used as source for TuDeT were carried out by foreigner research groups, most of the translations are in english, see Table~\ref{tab:source_link}. The lack of portuguese translations hinder their application for language learning purposes targeting brazilian people. In fact, we only had portuguese translations for ``Guajajara'', ``Munduruku'' and ``Akuntsu'' out of the 8 languages available. Additionally, we applied a data cleaning process to fix spelling and remove citations in the sentences themselves. Often the translated sentence would include a citation to its original work. For example, the sentence ``opo?o aka?u a?e'' was escorted by its portuguese translation ``ele colhe cacau (harrison, 2013:12)'', where there is a citation to the Portuguese-Guajajara dictionary~\cite{harrison2013dicionario}. The link between the treebank and TuLeD also face issues related to coverage, see Table~\ref{tab:source_link}. Finally, it is worth to note that the material available in the treebank was not necessarily designed for educational purposes and, thus, require moderation if ever properly applied in practice.

Besides the practical limitations of our data sources, it is worth to comment on some ethical concerns. BILingo prototype and its underlying data sources were designed without substantial indigenous community involvement~\cite{ijcai2023p685}. First, since TuLeD and TuDeT compile plenty sources from the literacy, it is hard to ensure that their data gathering procedures were compliant with ethical guidelines~\cite{indigenousprotocol20}, for example Los Pinos Declaration~\footnote{https://unesdoc.unesco.org/ark:/48223/pf0000374030}, or even if all their translations are validated by actual indigenous speakers. Here we should note that any language learning tool on BIL should rely on data sources that were carefully designed in partnership with indigenous communities.

\section{Conclusion} \label{sec:conclusion}

In this case of study, we explored the development of a language learning tool for BIL. We described how such a tool could work by detailing the student progression through course sections, lessons and exercises. We managed to use an existing dependency treebank (TuDeT) and a lexical database (TuLeD) to generate exercises. We were able to produce a working prototype and validate the potential of using dependency trees associated with lexical database to automatically generate exercises. Finally, we discussed the challenges and limitations of such system from practical and ethical perspectives.

Future work should develop data gathering protocols for creating treebanks and lexical databases with indigenous communities, and oriented for educational purposes so that we can have sufficient and reliable data sources to build effective learning tools in practice. Additionally, once it is possible to release such a system, research should be conducted to evaluate the engagement on indigenous communities and optimize the learning system so that students stay engaged. 

\bibliography{custom}

\begin{thebibliography}{16}
\expandafter\ifx\csname natexlab\endcsname\relax\def\natexlab#1{#1}\fi

\bibitem[{Akhlaghi et~al.(2019)Akhlaghi, Bédi, Butterweck, Chua, Gerlach, Habibi, Ikeda, Rayner, Sestigiani, and Zuckermann}]{akhlaghi19}
Elham Akhlaghi, Branislav Bédi, Matt Butterweck, Cathy Chua, Johanna Gerlach, Hanieh Habibi, Junta Ikeda, Manny Rayner, Sabina Sestigiani, and Ghil'ad Zuckermann. 2019.
\newblock \href {https://doi.org/10.21437/SLaTE.2019-19} {{Overview of LARA: A Learning and Reading Assistant}}.
\newblock In \emph{Proc. 8th ISCA Workshop on Speech and Language Technology in Education (SLaTE 2019)}, pages 99--103.

\bibitem[{B{\'e}di et~al.(2022)B{\'e}di, Beedar, Chiera, Ivanova, Maizonniaux, Chiar{\'a}in, Rayner, Sloan, and Zuckermann}]{branislav22}
Branislav B{\'e}di, Hakeem Beedar, Belinda Chiera, Nedelina Ivanova, Christele Maizonniaux, {Neasa N{\'i}} Chiar{\'a}in, Manny Rayner, John Sloan, and Ghil{\textquoteright}ad Zuckermann. 2022.
\newblock \href {https://doi.org/10.18653/v1/2022.computel-1.9} {Using lara to create image-based and phonetically annotated multimodal texts for endangered languages}.
\newblock In \emph{Proceedings of the Fifth Workshop on the Use of Computational Methods in the Study of Endangered Languages}, COMPUTEL 2022 - 5th Workshop on the Use of Computational Methods in the Study of Endangered Languages, Proceedings of the Workshop, pages 68--77. Association for Computational Linguistics (ACL).
\newblock Fifth Workshop on the Use of Computational Methods in the Study of Endangered Languages : Language Diversity: from Low-Resource to Endangered Languages, COMPUTEL-5 2022 ; Conference date: 26-05-2022 Through 27-05-2022.

\bibitem[{Gerardi et~al.(2022{\natexlab{a}})Gerardi, Reichert, Aragon, Martín-Rodríguez, Godoy, and Merzhevich}]{tudet22}
Fabrício~Ferraz Gerardi, Stanislav Reichert, Carolina Aragon, Lorena Martín-Rodríguez, Gustavo Godoy, and Tatiana Merzhevich. 2022{\natexlab{a}}.
\newblock \href {https://doi.org/10.5281/zenodo.6563353} {\emph{TuDeT: Tupían Dependency Treebank}}.
\newblock Zenodo.

\bibitem[{Gerardi et~al.(2022{\natexlab{b}})Gerardi, Reichert, Aragon, Wientzek, List, and Forkel}]{tuled22}
Fabrício~Ferraz Gerardi, Stanislav Reichert, Carolina Aragon, Tim Wientzek, Johann-Mattis List, and Robert Forkel. 2022{\natexlab{b}}.
\newblock \href {https://doi.org/10.5281/zenodo.6572576} {\emph{TuLeD. Tupían Lexical Database}}.
\newblock Zenodo.

\bibitem[{Harrison and Harrison(2013)}]{harrison2013dicionario}
Carl Harrison and Carole Harrison. 2013.
\newblock \emph{Dicionário Guajajara-Português}.
\newblock SIL.

\bibitem[{Katinskaia et~al.(2017)Katinskaia, Nouri, and Yangarber}]{katinskaia-etal-2017-revita}
Anisia Katinskaia, Javad Nouri, and Roman Yangarber. 2017.
\newblock \href {https://aclanthology.org/W17-0304} {{R}evita: a system for language learning and supporting endangered languages}.
\newblock In \emph{Proceedings of the joint workshop on {NLP} for Computer Assisted Language Learning and {NLP} for Language Acquisition}, pages 27--35, Gothenburg, Sweden. LiU Electronic Press.

\bibitem[{Lewis et~al.(2020)Lewis, Abdilla, Arista, Baker, Benesiinaabandan, Brown, Cheung, Coleman, Cordes, Davison, Duncan, Garzon, Harrell, Jones, Kealiikanakaoleohaililani, Kelleher, Kite, Lagon, Leigh, Levesque, Mahelona, Moses, Nahuewai, Noe, Olson, Parker~Jones, Running~Wolf, Running~Wolf, Silva, Fragnito, and Whaanga}]{indigenousprotocol20}
Jason~Edward Lewis, Angie Abdilla, Noelani Arista, Kaipulaumakaniolono Baker, Scott Benesiinaabandan, Michelle Brown, Melanie Cheung, Meredith Coleman, Ashley Cordes, Joel Davison, K{\=u}pono Duncan, Sergio Garzon, D.~Fox Harrell, Peter-Lucas Jones, Kekuhi Kealiikanakaoleohaililani, Megan Kelleher, Suzanne Kite, Olin Lagon, Jason Leigh, Maroussia Levesque, Keoni Mahelona, Caleb Moses, Isaac~('Ika'aka) Nahuewai, Kari Noe, Danielle Olson, '{\=O}iwi Parker~Jones, Caroline Running~Wolf, Michael Running~Wolf, Marlee Silva, Skawennati Fragnito, and H{\=e}mi Whaanga. 2020.
\newblock \href {https://spectrum.library.concordia.ca/id/eprint/986506/} {Indigenous protocol and artificial intelligence position paper}.
\newblock Project Report 10.11573/spectrum.library.concordia.ca.00986506, Aboriginal Territories in Cyberspace, Honolulu, HI.
\newblock Edited by Jason Edward Lewis. English Language Version of "Ka?ina Hana ?{\=O}iwi a me ka Waihona ?Ike Hakuhia Pepa K{\=u}lana" available at: https://spectrum.library.concordia.ca/id/eprint/990094/.

\bibitem[{Lightbown(2021)}]{lightbown21}
Spada~N. Lightbown, P.~M. 2021.
\newblock \emph{How Languages Are Learned (5 ed.)}.
\newblock Oxford University Press.

\bibitem[{Neubig et~al.(2020)Neubig, Rijhwani, Palmer, MacKenzie, Cruz, Li, Lee, Chaudhary, Gessler, Abney, Hayati, Anastasopoulos, Zamaraeva, Prud{'}hommeaux, Child, Child, Knowles, Moeller, Micher, Li, Zink, Xia, Sharma, and Littell}]{neubig-etal-2020-summary}
Graham Neubig, Shruti Rijhwani, Alexis Palmer, Jordan MacKenzie, Hilaria Cruz, Xinjian Li, Matthew Lee, Aditi Chaudhary, Luke Gessler, Steven Abney, Shirley~Anugrah Hayati, Antonios Anastasopoulos, Olga Zamaraeva, Emily Prud{'}hommeaux, Jennette Child, Sara Child, Rebecca Knowles, Sarah Moeller, Jeffrey Micher, Yiyuan Li, Sydney Zink, Mengzhou Xia, Roshan~S Sharma, and Patrick Littell. 2020.
\newblock \href {https://aclanthology.org/2020.sltu-1.48} {A summary of the first workshop on language technology for language documentation and revitalization}.
\newblock In \emph{Proceedings of the 1st Joint Workshop on Spoken Language Technologies for Under-resourced languages (SLTU) and Collaboration and Computing for Under-Resourced Languages (CCURL)}, pages 342--351, Marseille, France. European Language Resources association.

\bibitem[{Pinhanez et~al.(2023)Pinhanez, Cavalin, Vasconcelos, and Nogima}]{ijcai2023p685}
Claudio~S. Pinhanez, Paulo Cavalin, Marisa Vasconcelos, and Julio Nogima. 2023.
\newblock \href {https://doi.org/10.24963/ijcai.2023/685} {Balancing social impact, opportunities, and ethical constraints of using ai in the documentation and vitalization of indigenous languages}.
\newblock In \emph{Proceedings of the Thirty-Second International Joint Conference on Artificial Intelligence, {IJCAI-23}}, pages 6174--6182. International Joint Conferences on Artificial Intelligence Organization.
\newblock AI for Good.

\bibitem[{Rayner and Wilmoth(2023)}]{rayner-wilmoth-2023-using}
Manny Rayner and Sasha Wilmoth. 2023.
\newblock \href {https://aclanthology.org/2023.computel-1.3} {Using {LARA} to rescue a legacy {P}itjantjatjara course}.
\newblock In \emph{Proceedings of the Sixth Workshop on the Use of Computational Methods in the Study of Endangered Languages}, pages 13--18, Remote. Association for Computational Linguistics.

\bibitem[{Thomason(2015)}]{thomason15}
S.~G. Thomason. 2015.
\newblock \emph{Endangered Languages: An Introduction}.
\newblock Cambridge: Cambridge University Press.

\bibitem[{von Ahn(2006)}]{vonahn06}
L.~von Ahn. 2006.
\newblock \href {https://doi.org/10.1109/MC.2006.196} {Games with a purpose}.
\newblock \emph{Computer}, 39(6):92--94.

\bibitem[{Wurm(2001)}]{wurm01}
S.A. Wurm. 2001.
\newblock \emph{Atlas of the world’s languages in danger of disappearing}.
\newblock Unesco Pub.

\bibitem[{Xu et~al.(2022)Xu, U{\'\i}~Dhonnchadha, and Ward}]{xu-etal-2022-faoi}
Liang Xu, Elaine U{\'\i}~Dhonnchadha, and Monica Ward. 2022.
\newblock \href {https://doi.org/10.18653/v1/2022.computel-1.17} {Faoi gheasa an adaptive game for {I}rish language learning}.
\newblock In \emph{Proceedings of the Fifth Workshop on the Use of Computational Methods in the Study of Endangered Languages}, pages 133--138, Dublin, Ireland. Association for Computational Linguistics.

\bibitem[{Zuckermann et~al.(2021)Zuckermann, Vigf{\'u}sson, Rayner, N{\'\i}~Chiar{\'a}in, Ivanova, Habibi, and B{\'e}di}]{zuckermann-etal-2021-lara}
Ghil{'}Ad Zuckermann, Sigur{\dh}ur Vigf{\'u}sson, Manny Rayner, Neasa N{\'\i}~Chiar{\'a}in, Nedelina Ivanova, Hanieh Habibi, and Branislav B{\'e}di. 2021.
\newblock \href {https://aclanthology.org/2021.computel-1.3} {{LARA} in the service of revivalistics and documentary linguistics: Community engagement and endangered languages}.
\newblock In \emph{Proceedings of the 4th Workshop on the Use of Computational Methods in the Study of Endangered Languages Volume 1 (Papers)}, pages 13--23, Online. Association for Computational Linguistics.

\end{thebibliography}

\end{document}